\pdfoutput=1

\documentclass[11pt]{article}

\usepackage[]{acl}

\usepackage{times}
\usepackage{latexsym}

\usepackage[T1]{fontenc}
\usepackage{multirow}
\usepackage{times}
\usepackage{latexsym}
\usepackage{booktabs}
\usepackage[T1]{fontenc}
\usepackage{tipa}
\usepackage[utf8]{inputenc}

\usepackage{microtype}
\usepackage{inconsolata}

\usepackage{todonotes, color}
\newcommand{\Note}[2]{} 
\newcommand{\SideNote}[2]{}
\renewcommand{\Note}[2]{\todo[color=#1,size=\small, inline=true]{#2}} 
\renewcommand{\SideNote}[2]{\todo[color=#1,size=\small]{#2}} %

\usepackage{makecell}
\usepackage{xspace}
\newcommand{\std}[1]{{\scriptsize $\pm$#1}}
\newcommand{\base}{\textsc{base}\xspace}
\newcommand{\labreg}{\textsc{label}$_{reg}$\xspace}
\newcommand{\labvc}{\textsc{label}$_{vc}$\xspace}
\newcommand{\labboth}{\textsc{label}$_{2}$\xspace}

\newcommand{\typereg}{\textsc{type}$_{reg}$\xspace}
\newcommand{\typeirr}{\textsc{type}$_{irr}$\xspace}
\newcommand{\tokenreg}{\textsc{token}$_{both}$\xspace}
\newcommand{\tokenirr}{\textsc{token}$_{irr}$\xspace}

\newcommand{\balance}{\textsc{balance}\xspace}
\newcommand{\regds}{\textsc{reg}$_{ds}$\xspace}
\newcommand{\irrds}{\textsc{irreg}$_{ds}$\xspace}

\title{How do we get there? Evaluating transformer neural networks as cognitive models for English past tense inflection}

\author{Xiaomeng Ma \\
  The Graduate Center, CUNY\\
  \texttt{xma3@gradcenter.cuny.edu} \\\And
  Lingyu Gao \\
  Toyota Technological Institute at Chicago \\
  \texttt{lygao@ttic.edu} \\}

\begin{document}
\maketitle
\begin{abstract}
There is an ongoing debate on whether neural networks can grasp the quasi-regularities in languages like humans. In a typical quasi-regularity task, English past tense inflections, the neural network model has long been criticized that it learns only to generalize the \textit{most frequent} pattern, but not the \textit{regular} pattern, thus can not learn the abstract categories of regular and irregular and is dissimilar to human performance. In this work, we train a set of transformer models with different settings to examine their behavior on this task. The models achieved high accuracy on unseen regular verbs and some accuracy on unseen irregular verbs. The models' performance on the regulars is heavily affected by type frequency and ratio but not token frequency and ratio, and vice versa for the irregulars. The different behaviors on the regulars and irregulars suggest that the models have some degree of symbolic learning on the regularity of the verbs. In addition, the models are weakly correlated with human behavior on nonce verbs. Although the transformer model exhibits some level of learning on the abstract category of verb regularity, its performance does not fit human data well, suggesting that it might not be a good cognitive model. \footnote{The code and data for this paper can be found at: \href{https://github.com/xiaomeng-ma/English-Past-Tense}{https://github.com/xiaomeng-ma/English-Past-Tense.}} 

\end{abstract}

\section{Introduction}

Many aspects of language can be characterized as quasi-regular: the relationship between inputs and outputs is systematic but allow many exceptions. English past tense inflection exhibits such quasi-regularity that the regular verbs follow the `-\textit{ed}' rule (\textit{help} - \textit{helped}) and the irregular forms consist of a variety of changes such as changing vowel (\textit{sing} - \textit{sang}). There has been heated debate about how people represent regular and irregular for the past 40 years. For the single-route approach, \citet{100} described a feed-forward connectionist neural model that learned both regular and irregular forms of the English verbs' past tense without explicit symbolic rules. However, this model received fierce criticisms from the proponents of the dual-route model \cite[e.g.,][]{pinker1988language, marcus1992overregularization}, who argue that the speakers first reason over the abstract categories (regular - irregular), and process the regulars through rule-applying mechanism (adding -\textit{ed}) and process the irregulars via gradient analogical processes. In addition, \citet{pinker1988language} highlighted many empirical inadequacies of the model and argued that these failures stemmed from `central features of connectionist ideology' and would persist in any neural network model.

With the advancement of deep learning in NLP, there has been renewed interest in the English past tense debate with modern neural networks. \citet{kirov2018recurrent} revisited the past tense debate and showed that modern recurrent encoder-decoder (RNN) neural models overcame many of the criticisms. Their model achieved near-perfect accuracy on the unseen regular verbs and some accuracy on the unseen irregular verbs (28.6\% as 5 correct irregular verbs). In addition, the model's results on the nonce verb inflections correlate with human experimental data (Spearman's $\rho$ = 0.48 for regulars and $\rho$ = 0.45 for irregulars). Thus they concluded that the neural model could be a cognitive model. However, other studies have shown that the modern neural network is still susceptible to the criticism raised by \citet{marcus1995german}: the neural models lack symbolic rule learning ability and are vulnerable to the frequency distribution of the data, so they may learn to extend the \textit{most frequent} pattern, instead of the \textit{regular} pattern.  \citet{corkery2019we} closely examined the model's performance on the nonce verbs and found that the fit to the human data is weak, especially for the irregular verbs. Similarly, \citet{mccurdy2020inflecting} used German plural to demonstrate that the RNNs tend to overextend the most frequent plural class to nonce words and do not match the human speakers' data.  \citet{beser2021falling} found that in English and German plurals, transformers are also susceptible to the frequency distribution of the data as RNNs. Prior work has generally focused on the comparison between model's performance and human behavior on nonce verbs, and few have explored the neural model's behavior on English regular and irregular verbs. 

In our study, we closely examine the transformer's behavior on English past tense inflections corresponding to the training data's regular-irregular type and token frequency distributions to explore whether the models learn and apply symbolic rules. We train a set of transformers with different frequency distributions and experiment with resampling the training data for each epoch (\S\ref{sec:experiment}).    %
On our evaluation (\S\ref{sec:results}) of English verbs, the transformers achieved over 95\% accuracy on unseen regulars and some accuracy on unseen irregulars (ranging from 0\% - 22\%). We find that models exhibit different behaviors on the regulars and the irregulars, that the performance on regulars is more affected by the type frequency but not token frequency, and vice versa for the irregulars, suggesting that the models have some degree of abstract representation of verb regularity. We observe that the majority of the errors can be attributed to misclassification (e.g., treating an irregular as regular), with a smaller proportion of errors caused by applying the wrong inflection. 
For nonce verb evaluation (\S\ref{sec:nonce}), the models vary in correlations with human data. Generally, the models correlate with human data better on regulars than irregulars, but the overall correlations are weak. In conclusion, we found that the transformer models display some degree of abstract representation of verb regularity, but do not fit human data well, thus can not be a good cognitive model.

\section{Hypotheses and Predictions}
\subsection{Hypotheses}
We aim to investigate the transformer's ability to generalize symbolic categories and rules in English past tense inflection task. \citet{wei2021frequency} proposed three hypotheses for how a neural network processes the symbolic rules by analyzing the behavior of BERT model \cite{devlin-etal-2019-bert} on subject-verb agreement in English. We adapted their hypotheses and combined the theories of past tense debate to form our hypotheses. \textbf{H1: Idealized Symbolic Learner} operates over abstract categories and rules. For example, if x is a \textsc{regular} verb and x ends with /d/ or /t/, then \textsc{past}(x) = x + /\textsci d/. This is also the hypothesis for how humans process the regulars in the dual-route model. Under this hypothesis, the model would not misclassify verbs and is only sensitive to the type frequency, but not token frequency\footnote{\citet{wei2021frequency} suggested that the `idealized symbolic leaner would not be affected by word specific properties such as frequency', which we interpret as token frequency. In addition, psycholinguistic studies also suggested that human learners generalize phonological patterns based on type frequency and ignore the token frequency \cite[e.g.,][]{bybee2003phonology}}. \textbf{H2: Naive Pattern-Associating Learner} does not necessarily represent any abstract features of the input verbs (such as regular/irregular); instead, it produces the output by a neuron-like activation process, which is analogous to an early feed-forward network as proposed in \citet{100}. This is the foundation for modern transformers, because transformer models also incorporate feed-forward layers. Therefore, the transformer model would naturally fall under this hypothesis. \textbf{H3: Symbolic Learner with Noisy Observations} is a hybrid of H1 and H2, suggesting that the model at its core is a symbolic learner, but with noisy observations. The model is able to generalize the abstract category for regular and irregular verbs, as well as the inflection patterns. However, the noisy observations would affect its ability to map the inputs to the correct category and/or apply the appropriate past tense inflection. Under this hypothesis, the model's categorization ability is mainly affected by the type frequency, and the pattern generalization is affected by both type and token frequency. 

In this work, we expect the transformers to behave like H3, which operates based on pattern-associating and shows some level of symbolic learning. Moreover, the behavior on regular verbs should be a \textbf{\textsc{strong} Symbolic Learner with \textsc{less} noisy observations}, since the majority of English verbs are regular verbs and the regular inflection (adding /-d/, /-t/ or /\textsci d/) can be easily summarized as a rule. The behavior on the irregular verbs should be a \textbf{\textsc{weak} Symbolic Learner with \textsc{more} noisy observations}, given that there are less than 200 irregular verbs in English with many implicit irregular inflection patterns (e.g., \textit{go}-\textit{went}).   

\subsection{Predictions and Summary of Findings}
Since H2 is the basis of transformer models, we need to show that the model shows some symbolic learning ability to confirm H3. Evidence for symbolic learning includes type frequency effects and accurately classifying verbs into regulars and irregulars.  
In addition, we also need to demonstrate that the models exhibit stronger symbolic learning ability on regulars than irregulars. We would expect the regulars to display a strong type frequency effect and a weak token frequency effect, and vice versa for the irregulars. In addition, H3 learner predicts that the errors are due to failures to identify the verb as a regular verb, and/or apply the appropriate inflection. 

Our experiments (\S\ref{sec:results}) show that both regular and irregular verbs exhibit a clear type frequency effect and the models achieved good classification accuracy, suggesting some degree of symbolic learning. In addition, the regulars are more affected by the type frequency but not token frequency (and vice versa for the irregulars), suggesting that the regulars demonstrate stronger symbolic learning ability than the irregulars. The analysis also found misclassification errors and wrong inflection errors for the regulars and irregulars. 
\section{Data}
\label{sec:data}

The base dataset is the same one used in previous studies with English past tense, which includes 4,039 English verbs from the CELEX database \cite{baayen1995celex}. We converted the verbs to IPA symbols based on Carnegie Mellon University Pronouncing Dictionary using \texttt{eng-to-ipa} python package,\footnote{\href{https://pypi.org/project/eng-to-ipa/}{https://pypi.org/project/eng-to-ipa/}} and checked each verb's past tense forms on Merriam Webster dictionary.\footnote{\href{https://www.merriam-webster.com/}{https://www.merriam-webster.com/}} Among these verbs, 3,857 are regular verbs; 150 are irregular verbs; and 32 verbs have both regular and irregular forms, e.g., \textit{knit} - \textit{knit} or \textit{knitted}.\footnote{The counts are different from \citet{kirov2018recurrent} because the original dataset has some inconsistent labeling. Details are explained in Appendix.} We also created two labels for each verb: Regularity and Verb class. The regularity indicates whether the verb is regular or irregular. The verb class corresponds to the inflection of each verb, which includes three classes for regular verbs (/-d/, /-t/, /-\textsci d/) and seven classes for irregular verbs, including vowel change, vowel change +/-d/, vowel change +/-t/, ruckumlaut, weak, level and other \cite{cuskley2015adoption}. The examples for verbs of different regularities and verb class labels in the base dataset are shown in Table \ref{tab:base1}.\footnote{The 32 ambiguous verbs are treated as irregular in the table.}
\begin{table}[!t]
\small
\centering
\begin{tabular}{p{0.25\linewidth}p{0.25\linewidth}p{0.15\linewidth}p{0.15\linewidth}}
\toprule
 & Example & Count & \%\\
\midrule
\multicolumn{2}{l}{\textbf{Regular}} & 3857 & 95.5 \\
\midrule
/-d/ & called & 2045 & 50.6 \\
/-t/ & worked & 763 & 18.9 \\
/-\textsci d/ & wanted & 1049 & 26.0 \\
\midrule
\multicolumn{2}{l}{\textbf{Irregular}} & 182 & 4.5 \\
\midrule
vc& hide-hid & 125 & 3.1 \\
vc+/-t/ & feel-felt & 12 & 0.3 \\
vc+/-d/ & tell-told & 10 & 0.2 \\
ruck & buy-bought & 8 & 0.2 \\
weak & send-sent & 9 & 0.2 \\
level & quit-quit & 11 & 0.3 \\
other & go-went & 7 & 0.2\\
\bottomrule
\multicolumn{4}{l}{vc = vowel change, ruck = ruckumlaut}
\end{tabular}
\caption{The regularity and verb class distribution in the CELEX dataset (the ambiguous verbs are treated as irregulars).
}
\label{tab:base1}
\end{table}

\subsection{Test Data}
We evaluated the models on two test datasets: nonce verbs and real English verbs. Following the previous studies, we used 58 nonce verbs in \citet{albright2003rules} for comparison with human behavior. For the real English verb test dataset, we randomly selected 80 verbs from the CELEX database, including 60 regular verbs (20 per verb class) and 20 irregular verbs (2 verbs from vowel change + /-t/ class and 3 verbs from other classes). 

\subsection{Training Data}
After excluding the verbs in the test data, we developed 4 training datasets based on type frequency and token frequency. In the type frequency based training datasets, each verb appears only once.
Since there are 32 ambiguous verbs, we create \typereg where these verbs are all treated as regular, and \typeirr where they are all treated as irregular. 

Then we created \tokenreg, a token frequency based dataset with each verb appearing based on its CELEX frequency, where `both' indicates that we consider both regular and irregular forms for ambiguous words. For example, the irregular form \textit{knit} appears 5 times, and regular form \textit{knitted} appears 12 times. As regular verbs dominate all these 3 datasets, we created \tokenirr, where only the irregular verbs appear based on their CELEX frequency, and the regular verbs all appear once, of which the irregular rate is 92.3\%. The regular and irregular rates for all training sets are shown in Table \ref{tab:train}.

\begin{table}[!t]
\small
\centering
\begin{tabular}{p{0.13\linewidth}p{0.16\linewidth}p{0.14\linewidth}p{0.14\linewidth}p{0.14\linewidth}}
\toprule
\multicolumn{2}{l}{Training set}  & Regular & Irregular & {\begin{tabular}[c]{@{}c@{}}Total\\ tokens\end{tabular}} \\
\midrule
\multirow{2}{*}{\begin{tabular}[c]{@{}c@{}}Type\\ based\end{tabular}} & \typereg & 96.6\% & 3.4\%  & 3,959 \\
 & \typeirr & 95.9\%  & 4.1\%  & 3,959 \\
 \midrule
\multirow{2}{*}{\begin{tabular}[c]{@{}c@{}}Token\\ based\end{tabular}} & \tokenreg & 68.7\%  & 31.3\%  & 147,711 \\
 & \tokenirr & 7.7\%  & 92.3\%  & 49,983 \\
 \bottomrule
\end{tabular}
\caption{Regular and irregular verb distribution in different training datasets.}
\label{tab:train}
\end{table}
\section{Experiment}
\label{sec:experiment}
\subsection{Transformer Models}
We used the sequence-to-sequence transformers \cite{vaswani2017attention} to generate the past tense of the root verbs trained from scratch. Our \base model used the IPA phonemes of the root verb to generate the past tense inflections. We further examined whether identifying the regularity and verb class before generating the past tense would improve the model's performance. We added \labreg for regularity, \labvc for verb class, and \labboth for both. Examples of input and gold output in the training data are shown in Table \ref{tab:inputexample}. 

Since there are less than 200 irregular verbs in English, the model will be inevitably biased towards the regulars on type-based datasets. To adjust this imbalanced distribution, we downsample the number of regular verbs to match the number of irregulars in training data per epoch on \typeirr, which we called \balance.\footnote{There are 162 irregular verbs (excluding 20 verbs in test) in \typeirr. The train-dev split is 80-20, yielding 129 irregular verbs in training. We choose \typeirr as it contains the most number of unique irregulars.} To investigate the type-frequency effect, we further apply two unbalanced resampling methods per epoch:\footnote{We keep the numbers of irregular verbs unchanged, as we would prefer the model to see all irregular verbs for higher accuracy on irregulars.} \regds downsizes the regulars to match the decreased regular rate in
Parents' Data.\footnote{We selected 8 children's corpora in the CHILDES database \cite{macwhinney2000childes} and aggregated their parents' past tense verbs. If we leverage the percentage of its irregulars with the same construction method of \tokenirr, the irregular rate is 72.6\%. Details are shown in Appendix \ref{sec:appendix:parent}.}
, and \irrds downsizes the irregulars to match the irregular rate in \tokenirr. Count of regular and irregular verbs, as well as irregular ratio seen per training epoch are listed in Table \ref{tab:epoch}.

In addition, we added a pointer-generator mechanism \cite{vinyals2015pointer} to the transformer model to reduce bizarre errors like *\textit{membled} for \textit{mailed} that was reported in \citet{100}'s original model\footnote{\citet{kirov2018recurrent} also reported one instance of this type of error and suggested that this type of errors could be eliminated by increasing training epochs. This type of errors has also been reported in other inflection tasks such as text normalization \cite{zhang2019neural}.}. This model could choose between generating a new element and copying an element from the input directly to the output. Transformers with copy mechanism have been used for word-level tasks \cite{zhao2019improving} and character-level inflections \cite{singer2020nyu}. 

\begin{table}[!t]
\small
\centering
\begin{tabular}{p{0.2\linewidth}p{0.6\linewidth}}
\toprule
Input & Start, k, \textopeno, l, End \\
\midrule
Model & Output \\
\midrule
\base & Start, k, \textopeno, l, d, End \\
\labreg & Start, reg, k, \textopeno, l, d, End \\
\labvc & Start, +d, k, \textopeno, l, d, End \\
\labboth & Start, reg, +d, k, \textopeno, l, d, End \\
\bottomrule
\end{tabular}
\caption{Input and gold output in the training data with different labels for the verb `call', tokens are separated by comma.}
\label{tab:inputexample}
\end{table}

\begin{table}[!t]
\small
\centering
\begin{tabular}{p{0.17\linewidth}p{0.14\linewidth}p{0.14\linewidth}p{0.25\linewidth}}
\toprule
 Resample & Count\textsubscript{Reg} & Count\textsubscript{Irr} & Irr. ratio (\%) \\
 \midrule
\balance & 129 & 129 & 50.0 \\
\regds & 48 & 129 & 72.6 \\
\irrds & 283 & 129 & 31.3\\
\bottomrule
\end{tabular}
\caption{Count of regular (Reg) and irregular (Irr) verbs in three epoch training datasets. Irr. ratio denotes the percentage of irregular verbs in training data per epoch.}
\label{tab:epoch}
\end{table}

\subsection{Experiment Setups}
Both encoder and decoder of our models have 2 layers, 4 attention heads, 128 expected features in the input, and 512 as the dimension of the feed-forward network model. For training, we split the dataset into train-dev splits of 90-10, set model dropout to 0.1, and used Adam optimizer \cite{kingma2014adam} with varied learning rate in the training process computed according to \citet{vaswani2017attention}. Besides, we set batch size to 32 for type-based datasets, 64 for \tokenirr, and 128 for \tokenreg. We run 30 epochs for all datasets. When we apply resampling methods (\balance, \regds, and \irrds), we set batch size to 8 and run 100 epochs, as there's fewer data per training epoch. As most of the datasets are highly unbalanced, we compute accuracy for both regular verbs and irregular verbs on dev set, and average them to select the best model.  
For inference, we set beam size to 5. 

\section{Results}
\subsection{English verbs' Test Accuracy}
\label{sec:results}
\begin{table}[!t]
\small
\centering
\begin{tabular}{llllll}
\toprule
\multirow{2}{*}{Train Set}& \multirow{2}{*}{Model} &\multicolumn{2}{c}{Regular}&\multicolumn{2}{c}{Irregular}\\
& &van.&copy&van.&copy\\
\midrule
\typereg&\base&99.0&99.0&\textbf{4.0}&0.0\\
&\labreg&97.3&\textbf{99.7}&0.0&\textbf{1.0}\\
&\labvc&\textbf{99.3}&98.3&1.0&\textbf{1.0}\\
&\labboth&99.0&\textbf{99.7}&1.0&0.0\\
\midrule
\typeirr&\base&97.0&97.0&\textbf{2.0}&\textbf{3.0}\\
&\labreg&\textbf{99.0}&\textbf{99.7}&0.0&1.0\\
&\labvc&94.7&99.3&0.0&0.0\\
&\labboth&97.0&97.7&0.0&1.0\\
\midrule
\tokenreg&\base&\textbf{98.0}&\textbf{99.3}&\textbf{11.0}&\textbf{8.0}\\
&\labreg&96.7&97.0&10.0&4.0\\
&\labvc&97.7&97.0&2.0&2.0\\
&\labboth&\textbf{98.0}&97.0&4.0&3.0\\
\midrule
\tokenirr&\base&\textbf{95.7}&96.0&\textbf{22.0}&4.0\\
&\labreg&95.0&\textbf{97.7}&9.0&\textbf{12.0}\\
&\labvc&93.0&96.3&5.0&10.0\\
&\labboth&95.0&94.3&6.0&5.0\\
\bottomrule
\end{tabular}
\caption{Test accuracy (\%) for our models for regular and irregular verbs, where `van.' and `copy' refer to the vanilla transformer model and the transformer model with pointer-generator mechanism respectively.}
\label{tab:en_test_acc}
\end{table}
We calculated the test accuracy of our models based on the regulars and irregulars in the real English verb test set, which is shown in Table~\ref{tab:en_test_acc}.\footnote{All accuracy in this paper are averaged over 5 runs with different random seeds, while errors are counted by summing up the errors of different runs.} For all models, the regular verbs' accuracy was over 93\%,  and the irregular accuracy ranges from 0\%-22\% where the token-based models have better accuracy. The copy mechanism improved the accuracy for regular verbs, as we expected. The \labreg, \labvc, and \labboth did not improve the irregulars accuracy for the vanilla model. The accuracy for each verb class can be found in Appendix Table~\ref{tab_app:vc_regular} and Table~\ref{tab_app:vc_irregular}.

\paragraph{Testing H1: Evidence for Symbolic Learning}
\begin{table}[t]
\small
\centering
\begin{tabular}{llllll}
\toprule
\multirow{2}{*}{Test Acc} & \multirow{2}{*}{Model} &\multicolumn{2}{c}{Regular}&\multicolumn{2}{c}{Irregular}\\
& &van.&copy&van.&copy\\
\midrule
\balance&\base&72.7&\textbf{74.7}&23.0&\textbf{24.0}\\
\scriptsize{irr:129}&\labreg&71.0&62.3&\textbf{24.0}&21.0\\
\scriptsize{reg:129} &\labvc&68.7&71.3&17.0&18.0\\
&\labboth&\textbf{74.0}&68.7&19.0&14.0\\
\midrule
\regds&\base&\textbf{58.7}&\textbf{61.3}&\textbf{32.0}&25.0\\
\scriptsize{irr:129}&\labreg&56.7&52.7&23.0&\textbf{28.0}\\
\scriptsize{reg:48}&\labvc&56.0&52.0&21.0&20.0\\
&\labboth&55.7&60.3&21.0&15.0\\
\midrule
\irrds&\base&77.0&\textbf{85.3}&\textbf{21.0}&15.0\\
\scriptsize{irr:129}&\labreg&82.3&73.7&16.0&15.0\\
\scriptsize{reg:283}&\labvc&\textbf{83.3}&72.7&14.0&\textbf{16.0}\\
&\labboth&79.7&81.7&12.0&10.0\\
\bottomrule
\end{tabular}
\caption{Test accuracy (\%) for models trained on resampled data of \typeirr, where van. refers to vanilla model without copy mechanism. The irregular and regular tokens per epoch are listed for each resampling method.} 
\label{tab:test_acc_resample}
\end{table}

\begin{table}[]
\centering
\small
\begin{tabular}{llllll}
\toprule
\multirow{2}{*}{Label Acc} & \multirow{2}{*}{Model} & \multicolumn{2}{c}{Regular} & \multicolumn{2}{c}{Irregular} \\
 &  & van. & copy & van. & copy \\
 \midrule
\balance & \labreg & 77.3 & 72.3 & \textbf{79.0} & \textbf{85.0} \\
 & \labboth & \textbf{85.3} & \textbf{83.7} & 61.0 & 72.0 \\
 \midrule
\regds & \labreg & 60.7 & \textbf{72.0} & \textbf{90.0} & \textbf{88.0} \\
 & \labboth & \textbf{66.0} & 65.3 & 87.0 & 82.0 \\
 \midrule
\irrds & \labreg & \textbf{90.0} & 82.0 & 54.0 & \textbf{59.0} \\
 & \labboth & 85.3 & \textbf{87.7} & \textbf{55.0} & 55.0\\
 \bottomrule
\end{tabular}
\caption{Regularity label accuracy (\%) for models with different resampled methods.}
\label{tab:label_acc_resample}
\end{table}
To show that the models exhibit some level of symbolic learning, we first examine the test accuracy of resampling method to explore the type frequency effect. As shown in Table~\ref{tab:test_acc_resample}, the accuracies of the regular verbs increase as their type frequency and ratio increase, showing the type frequency effect. In addition, the irregular verbs exhibit a relative type frequency effect too, that the accuracy increases as the type ratio increases, while the absolute frequency remains the same. 

We further calculated the regularity label's accuracy on \labreg and \labboth to examine the model's ability to categorize verbs into regulars and irregulars. As shown in Table~\ref{tab:label_acc_resample}, the models achieved good label accuracy for both regulars and irregulars, suggesting that the model has the ability to correctly classify the verbs. The label accuracies also display a type frequency effect, that the accuracies increased as the type frequency and ratio increased. These findings confirm that the model exhibits some level of symbolic learning. 

\paragraph{Regular vs Irregular: Strong vs Weak Symbolic Learner}  We first examine the type and token frequency effect on the regulars and irregulars. The regular accuracy should be affected more by the type frequency than the token frequency, and vice versa for the irregulars. For the type frequency effect, we calculated the accuracy change for different models of \typereg and \typeirr in Table~\ref{tab:en_test_acc}. The regular's accuracies are more affected by the change of type frequency than the irregulars, with higher average change and max change, as listed in Table~\ref{tab:aac}. For token frequency effect, we calculated the accuracy change in \typeirr and \tokenirr where the regular and irregular's type frequency remains the same, but token frequency increased in both training datasets. The irregulars are more affected by the change of token frequency than the regulars, as listed in Table~\ref{tab:aac}. 

\begin{table}[]
\small
\centering
\begin{tabular}{llll}
\toprule
\multicolumn{2}{l}{Accuracy change} & mean\std{std}  & max \\
 \midrule
Type Freq. Effect& reg & 1.2\std{2.0} & 4.7 \\
\scriptsize{\typereg - \typeirr}  & irreg & 0.1\std{1.6} & -3.0 \\
\midrule
Token Freq. Effect  & reg & 0.1\std{2.2} & -3.0 \\
\scriptsize{\typeirr - \tokenirr} & irreg & -4.6\std{3.2} & -10.0 \\
\bottomrule
\end{tabular}
\caption{The accuracy change (\%) for type frequency effect comparison (\typereg - \typeirr) and token frequency comparison (\typeirr - \typereg).}
\label{tab:aac}
\end{table}
Next, we examine the model's classification ability. We manipulate the inferencing process for \labreg and \labboth models by manually setting the regularity label to the gold label\footnote{For example, for the verb \textit{rethink}, the \labreg will first output the `reg' or `irreg' label before producing the past tense. We manually set the label to `irreg' and let the model predict based on the set label.} and let the model output the past tense based on the correct category. This method allows us to explore how classification affects test accuracy. The accuracy results for different models after inferencing is listed in Table~\ref{tab:inference}. Inferencing improved the accuracy for the irregulars more than the regulars. This result indicates that misclassification errors are frequent for irregulars, but not regulars, suggesting that the models have a stronger classification ability for the regulars than the irregulars. 

In summary, the transformers exhibit stronger symbolic learning ability on the regulars than the irregulars that regular accuracy is more affected by type frequency but not token frequency, and vice versa for the irregulars. The models made fewer errors due to classification on the regulars than the irregulars. 

\begin{table}[]
\small
\centering
\begin{tabular}{llllll}
\toprule
\multirow{2}{*}{Train Set} & \multirow{2}{*}{Model} & \multicolumn{2}{c}{Regular} & \multicolumn{2}{c}{Irregular} \\
 &  & van & copy & van & copy \\
\midrule
\typereg & \labreg & 98.7 & 99.7\# & 22.0 & 14.0 \\
 & \labboth & 99.0\# & 100.0 & 36.0 & 24.0 \\
 \midrule
\typeirr & \labreg & 99.0\# & 99.7\# & 29.0 & 21.0 \\
 & \labboth & 98.3 & 99.0 & 39.0 & 32.0 \\
 \midrule
\tokenreg & \labreg & 99.0 & 99.7 & 54.0 & 31.0 \\
 & \labboth & 99.7 & 100.0 & 56.0 & 53.0 \\
 \midrule
\tokenirr & \labreg & 98.3 & 100.0 & 50.0 & 30.0 \\
 & \labboth & 99.0 & 99.3 & 48.0 & 57.0 \\
 \midrule
\end{tabular}
\caption{Test accuracy (\%) after inferencing by setting the regularity label to the gold label. \# indicates no change compared to the test accuracy without inferencing in Table~\ref{tab:en_test_acc}.}
\label{tab:inference}
\end{table}

\subsection{Error Analysis}
We further conduct error analysis on regular and irregular verbs. H3 predicts the model to make classification errors as well as inflection pattern errors. The regulars should have a lower percentage of both types of errors than the irregulars, since it is a \textsc{stronger} symbolic learner with less noisy observations. 

We categorized the regular and irregular errors into classes based on the H3's prediction: \textbf{1. classification errors}, where the model output an irregular form for a regular verb, or a regular form for the irregular, \textbf{2. inflection errors} where the model applied a wrong regular inflection to a regular verb or a wrong irregular inflection to an irregular verb. In addition, for regular verbs, we also found \textbf{copy errors} where the model copied the verb root incorrectly, and \textbf{creative errors} for the irregulars where the model output some unseen inflection patterns. All errors of the models in Table~\ref{tab:en_test_acc} are manually annotated by researchers with linguistic training. The counts and examples for each error type are listed in Table~\ref{tab:en_error_type}. The proportions of classification and inflection errors are lower for the regulars than the irregulars, further providing evidence for regular as \textsc{strong} symbolic learner. 

\begin{table}[!t]
\small
\centering
\begin{tabular}{p{0.3\linewidth}p{0.25\linewidth}p{0.25\linewidth}}
\toprule
Regular Error & Counts & Example \\
\midrule
classification & 144 (57.3\%) & fine: /fa\textupsilon n/ \\
inflection & 15 (6.0\%) & coach: /ko\textupsilon \textteshlig d/ \\
copy & 92 (36.7\%) & unleash: /\textschwa ni\textesh t/ \\
\midrule
Irregular Error & Counts & Example \\ 
\midrule
classification & 2755 (89.8\%) &  seek: /sikt/\\
inflection & 279 (9.1\%) & abide: /\textschwa ba\textupsilon d/  \\
creative & 34 (1.1\%) & forgo: /f\textopeno rgru/\\
\bottomrule
\end{tabular}
\caption{The counts and examples of regular error types and irregular error types. Counts are computed by summing up errors of all the models listed in Table~\ref{tab:en_test_acc}.}
\label{tab:en_error_type}
\end{table}
We further examined the copy errors for the regular verbs. Most of the errors either omit a consonant if two consonants are next to each other, e.g., \textit{un\textbf{l}eash}: /\textschwa ni\textesh t/, \textit{hitch\textbf{h}ike}: /h\textsci \textteshlig a\textsci kt/, or omitting a vowel if two vowels appear adjacent, e.g., \textit{tri\textbf{u}mph}: /tra\textsci mft/, \textit{co-\textbf{o}pt}: /ko\textupsilon pt\textsci d/. This pattern suggests that the models might have learned that consonant or vowel clusters are not likely to appear in English, thus adjusting its output to avoid improbable consonant and vowel clusters. 
\subsection{Nonce verbs' correlation with humans}
\label{sec:nonce}
In this section, we compared the models' performance with human behavior by correlating the results on nonce verbs. The human experiment data is from two experiments run by \citet{albright2003rules}. They created 58 nonce English verbs and assigned regular and irregular past tense forms to each verb, e.g., \textit{bize}: /ba\textsci zd/, /bo\textupsilon z/. 16 of these verbs were assigned 2 irregular forms, e.g., \textit{rife}: /ro\textupsilon f/ and /r\textsci f/. The participants were asked to first produce the past tense forms of these verbs, resulting in a production probability ($P_{pro}$), and to rate the regular and irregular forms of the past tense verbs, yielding a rating score. We follow \citet{corkery2019we}'s practice by treating each model as an individual participant and using the aggregated results to compare with the human results. To calculate the model's production probability, we used top-k sampling method to generate the top 5 outputs for each nonce verb, and aggregated the results over 5 random seeds. The model's production probability of each verb form is aggregated over 25 outputs. We correlated the model's $P_{pro}$ with human's $P_{pro}$ using Pearson $r$ and used Spearman $\rho$ to correlated the model's $P_{pro}$ and humans' rating score. 

The correlations with human data vary a lot among our models with different settings, i.e., some models could achieve a correlation over 0.7, while other models have negative correlations with human's data. The summary of the correlations' statistics of all the models is listed in Table \ref{tab:cormean}. Detailed correlation for each model can be found in Table \ref{tab_app:cor} in Appendix. The \labvc + \tokenreg model (vanilla \labvc trained on \tokenreg) achieves the best overall correlation with human data, as is listed in Table~\ref{tab:correg}. This model has a higher correlation with regular verbs than irregular verbs. For the models trained on resampled data, the \base + \balance (vanilla \base model with \balance resampling method) achieved the best overall correlation, as listed in Table~\ref{tab:corbalance}.

\begin{table}[!t]
\small
\centering
\begin{tabular}{lllll}
\toprule
 &  & Mean & Std & Range \\
 \midrule
Regular & $P_{pro} r$ & 0.31 & 0.29 & [-0.19, 0.70] \\
 & Rate $\rho$ & 0.48 & 0.21 & [0.02, 0.79] \\
Irregular & $P_{pro} r$& 0.32 & 0.13 & [0.06, 0.62] \\
 & Rate $\rho$& 0.31 & 0.12 & [-0.06, 0.55] \\
Irregular 2 &$P_{pro} r$& 0.25 & 0.28 & [-0.25, 0.77] \\
 & Rate $\rho$ & 0.18 & 0.16 & [-0.25, 0.61] \\
 \bottomrule
\end{tabular}
\caption{The mean, standard deviation, and range for the correlation of different models (including all the models in Table~\ref{tab:en_test_acc} and the models in Table~\ref{tab:test_acc_resample}). Irregular 2 stands for the 16 verbs with 2 irregular forms. $P_{pro}$ represents the production probability.}
\label{tab:cormean}
\end{table}

\begin{table}[!t]
\small
\centering
\begin{tabular}{p{0.4\linewidth}p{0.2\linewidth}p{0.2\linewidth}}
\toprule
\labvc + \tokenreg & $P_{pro}$ ($r$) & Rating ($\rho$) \\
 \midrule
Regular (N = 58) & 0.57 & 0.59 \\
Irregular (N = 58) & 0.22 & 0.22 \\
Irregular 2 (N = 16) & 0.12 & 0.36 \\
\bottomrule
\end{tabular}
\caption{The correlations with human's data for vanilla \labvc trained on \tokenreg.}
\label{tab:correg}
\end{table}

\begin{table}[!t]
\small
\centering
\begin{tabular}{p{0.4\linewidth}p{0.2\linewidth}p{0.2\linewidth}}
\toprule
\base + \balance & $P_{pro}$ ($r$) & Rating ($\rho$) \\
 \midrule
Regular (N = 58) & 0.62 & 0.74 \\
Irregular (N = 58) & 0.44 & 0.45 \\
Irregular 2 (N = 16) & 0.69 & 0.28 \\
\bottomrule
\end{tabular}
\caption{The correlations with human's data for vanilla \base model with \balance resampling method.}
\label{tab:corbalance}
\end{table}
\begin{figure*}[!t]
    \centering
    \includegraphics[width = 0.9\linewidth]{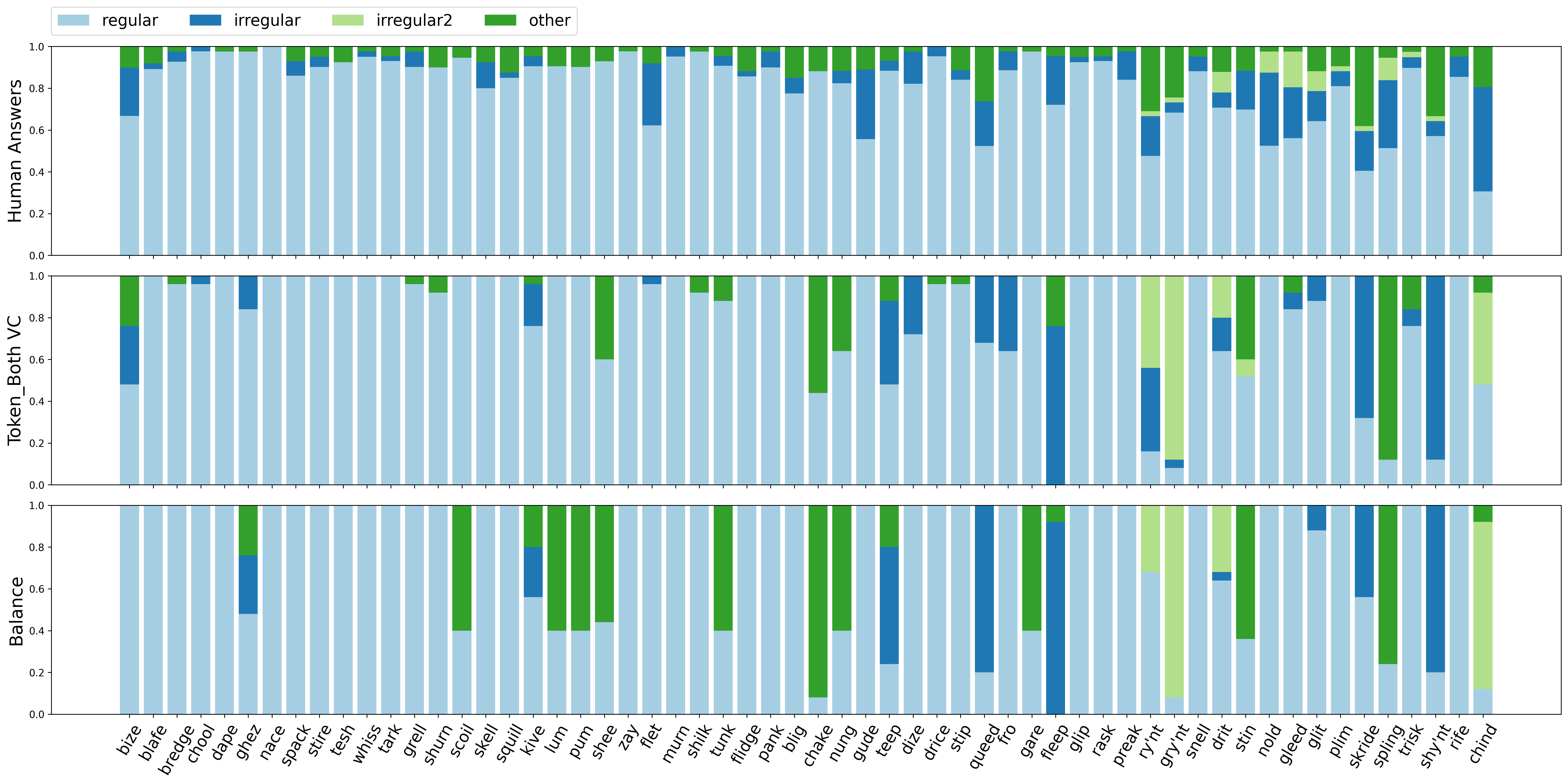}
    \caption{Percentage of regular, irregular, irregular 2 and other responses produced by humans (top), \labvc + \tokenreg model and \base+ \balance model. The last 16 verbs (starting with `preak') have 2 irregular forms.}
    \label{fig:correlation}
\end{figure*}
In addition, we plotted \labvc + \tokenreg, \base+ \balance and human's production probability for each nonce verb in Figure \ref{fig:correlation}. Human speakers are generally able to produce some irregular forms for the nonce verbs, except for only one verb (\textit{nace}). The models are less flexible in producing irregular forms. The \labvc + \tokenreg model only produced the regular forms for 27 verbs and 36 verbs for the \base+ \balance model. For the verbs with 2 irregular forms, humans are able to produce both forms for most of the verbs except for 3 verbs. However, the models' behaviors are more extreme that they are more likely to output only one type of irregular form of the verb. In addition, models and humans both produced many `other' forms that are not included in \citet{albright2003rules}. For models, the `other' forms are usually alternative irregular forms. For example, for the verb `shee' /\textesh i/, model's `other' output include /\textesh \textepsilon/, /\textesh \textopeno/, /\textesh it/.  Due to a lack of description of the `other' output in human data, we could not closely examine whether model's other outputs are similar to humans. 

In conclusion, it's difficult to make a simple statement whether the model behaves like the human. Our best-performing models are able to achieve a high correlation with regular verbs in human's data, but a weak correlation for irregular data. In addition, with a closer examination of the verb by verb production probability, it seems that humans are more flexible in generating regular or irregular verbs than the models. In human's data, although the regular form appears to be dominant for most of the verbs, the various irregulars can still be produced even with such strong regular preference. The models lack such flexibility and produce the outputs in a more absolute manner. For example, the models output only the regular forms of many verbs and do not output any irregular forms. Similarly, there are also verbs that the models produce the vast majority of irregular forms. The models are more strongly influenced by their regular or irregular bias on each verb than humans. 
\section{Discussion and Conclusion}
In this work, we demonstrate that the transformer models exhibit some abstract representation of regular and irregular verbs in past tense inflection generation. This abstract representation is largely affected by the type frequency of the input data. Since the regulars have a higher type frequency, the abstract representation is more robust for regular verbs than the irregular verbs. In addition, as long as the model could correctly classify the regular verb, it rarely makes errors in applying the correct inflection. Given the low type frequency and highly diverse inflection patterns for the irregular verbs, it is challenging for the model not only to classify the irregulars correctly, but also to apply the appropriate inflections. We found that increasing the type ratio would improve classification, and increasing token frequency would improve applying the correct inflections. 

In addition, we also compared the model's nonce verb output with human data. The correlation with human data varies greatly for different models, which makes it difficult to state whether the neural models can capture human behavior. In our best-performing model, we observe that the model is able to produce both regular and irregular forms for a nonce verb. However, the models are more influenced by  their own regular or irregular bias than human speakers. For example, humans can generate various forms even with a strong preference for regulars. However, the models are likely to generate either regular or irregular forms for a certain verb. Thus we conclude that the model's performance does not fit human data well. 

Neural models have long been viewed as an approach against abstract representations. Therefore, the neural models are often rejected as cognitive models. In our work, we showed that the models exhibit some abstract representations, although still have a weak correlation with human performance for different reasons. We hope our findings could imply that the dual-route mechanism is not necessarily against each other and lead to more discussions about incorporating both sides of the debate to build a better cognitive model.

\bibliography{anthology,custom}
\bibliographystyle{acl_natbib}

\clearpage

\section{Appendix}\label{sec:appendix}

\subsection{Data}
\subsubsection{Parents' Data}\label{sec:appendix:parent}
We created a dataset with parents' input past verbs with a higher irregular rate. We selected 8 children's corpora in the CHILDES database \cite{macwhinney2000childes} and aggregated their parents' past tense verbs. These 8 children include Adam, Eve, Sarah, Peter \cite{bloom1973one}, Allison \cite{bloom1974imitation}, Naomi \cite{sachs1983talking}, April \cite{higginson1985fixing}, and Fraser \cite{lieven2009two}. All 8 children have been extensively studied in the previous literature to show that they have overregularization errors at an early age. However, we didn't use it as one of our training sets, because this dataset is too small for training from scratch, including only 411 unique past tense verbs with 69 unique irregulars (irregular verb ratio is 16.8\%). If we leverage the percentage of its irregulars with the same construction method of \tokenirr, the dataset size would be 13,854 with an irregular ratio of 72.6\%, which we used for the irregular ratio for \regds.

\subsection{Data Cleaning}
We cleaned the dataset used in KC \cite{kirov2018recurrent} by checking each verb's past tense in Merriam Webster dictionary 
and annotating the pronunciation of each verb with IPA. In KC's dataset, 14 verbs' past tenses and their labels are inconsistent, which are labeled with * in Table \ref{tab_app:verb}, and 2 verbs' past tenses are inconsistent with Merriam Webster dictionary, which are labeled with $\dagger$. There are 33 verbs that have both regular and irregular past tense. 
\begin{table*}[]
\small
\centering
\begin{tabular}{llll}
\toprule
Verb & KC's past tense & KC's label & Merriam Webster \\
\midrule
\multicolumn{4}{l}{Verbs with both regular and irregular past tense} \\
\hline
abide & abided & reg & abided, abode \\
alight & alighted & reg & alighted, alit \\
awake & awoke & irreg & awoke, awaked \\
beseech & besought & irreg & beseeched, besought \\
bet & betted & irreg* & bet, betted \\
broadcast & broadcasted & reg & broadcast, broadcasted \\
cleave & cleaved & reg & cleaved, clove, clave \\
clothe & clothed & reg & clothed, clad \\
dive & dived & irreg* & dived, dove \\
dream & dreamed & irreg* & dreamed, dreamt \\
floodlight & floodlighted & reg & floodlit, floodlighted \\
gild & gilded & reg & giled, gilt \\
gird & girded & reg & girded, girt \\
hang & hung & irreg & hung, hanged \\
inset & insetted & irreg* & inset, insetted \\
knit & knitted & irreg* & knit, knitted \\
leap & leaped & irreg* & leaped, leapt \\
light & lighted & irreg* & lit, lighted \\
outshine & outshone & irreg & outshone, outshined \\
plead & pleaded & reg & pleaded, pled \\
quit & quitted & irreg* & quit, quitted \\
rend & rent & reg* & rent, rended \\
shine & shone & irreg & shone, shined \\
shoe & shod & reg* & shod, shoed \\
sneak & sneaked & irreg* & sneaked, snuck \\
speed & speeded & irreg* & sped, speeded \\
spit & spat & irreg & spit, spat, spitted \\
stick & stuck & irreg & sticked, stuck \\
strive & strove & irreg & strove, strived \\
sweat & sweated & reg & sweat, sweated \\
tread & trod & irreg & trod, treaded \\
wed & wedded & reg & wedded, wed \\
wet & wetted & irreg* & wet, wetted \\
\midrule
\multicolumn{4}{l}{Verbs with more than one irregular past tense.} \\
\hline
beget & begot & irreg & begot, begat \\
bid & bade & irreg & bade, bid \\
sing & sang & irreg & sing, sung \\
sink & sank & irreg & sank, sunk \\
\midrule
\multicolumn{4}{l}{KC's data inconsisted with Merriam Webster} \\
\hline
cost & costed$\dagger$ & irreg* & cost \\
shit & shitted$\dagger$ & reg & shit, shat\\
\bottomrule
\end{tabular}
\caption{The verbs and their past tense listed in KC's dataset and Merriam Webster dictionary. *indicates that the KC's label and its past tense do not match. $\dagger$ indicates the past tense in KC is not listed in the dictionary.}
\label{tab_app:verb}
\end{table*}

\subsection{Accuracy by Verb Class}
We report the test accuracy by verb class on regulars/irregulars of different models in Table~\ref{tab_app:vc_regular} and Table~\ref{tab_app:vc_irregular}.
\begin{table*}[]
\small
\centering
\begin{tabular}{llllllll}
\toprule
Train Set & Model & \multicolumn{2}{c}{/-d/} & \multicolumn{2}{c}{/-t/} & \multicolumn{2}{c}{/\textsci d/} \\
 &  & van. & copy & van. & copy & van. & copy \\
 \midrule
\typereg & \base & 100 & 100 & 94 & 98 & 94 & 97 \\
 & \labreg & 100 & 99 & 98 & 96 & 98 & 97 \\
 & \labvc & 99 & 100 & 97 & 97 & 97 & 97 \\
 & \labboth & 98 & 99 & 97 & 99 & 98 & 96 \\
 \midrule
\typeirr & \base & 99 & 100 & 96 & 98 & 92 & 98 \\
 & \labreg & 100 & 98 & 100 & 98 & 94 & 96 \\
 & \labvc & 98 & 99 & 95 & 94 & 99 & 98 \\
 & \labboth & 99 & 99 & 96 & 94 & 97 & 93 \\
 \midrule
\tokenreg & \base & 95 & 99 & 97 & 99 & 100 & 98 \\
 & \labreg & 96 & 96 & 98 & 97 & 99 & 100 \\
 & \labvc & 98 & 94 & 95 & 98 & 98 & 99 \\
 & \labboth & 98 & 95 & 99 & 98 & 97 & 100 \\
 \midrule
\tokenirr & \base & 95 & 95 & 93 & 99 & 98 & 98 \\
 & \labreg & 94 & 92 & 98 & 97 & 95 & 96 \\
 & \labvc & 94 & 90 & 96 & 95 & 97 & 96 \\
 & \labboth & 92 & 93 & 94 & 96 & 96 & 95 \\
 \bottomrule
\end{tabular}
\caption{Test accuracy (\%) of different models on regulars by verb class.}
\label{tab_app:vc_regular}
\end{table*}

\begin{table*}[]
\footnotesize
\centering
\begin{tabular}{llc@{\hspace{0.5ex}}cc@{\hspace{0.5ex}}cc@{\hspace{0.5ex}}cc@{\hspace{0.5ex}}cc@{\hspace{0.5ex}}cc@{\hspace{0.5ex}}cc@{\hspace{0.5ex}}c}
\toprule
 &  & \multicolumn{2}{c}{vc} & \multicolumn{2}{c}{vc+/-t/} & \multicolumn{2}{c}{vc+/-d/} & \multicolumn{2}{c}{ruck} & \multicolumn{2}{c}{weak} & \multicolumn{2}{c}{level} & \multicolumn{2}{c}{other} \\
 &  & van. & copy & van. & copy & van. & copy & van. & copy & van. & copy & van. & copy & van. & copy \\
 \midrule
\typereg & \base & 6.7 & 0 & 6.7 & 0 & 0 & 6.7 & 0 & 0 & 6.7 & 0 & 0 & 0 & 13.3 & 6.7 \\
 & \labreg & 0 & 0 & 0 & 0 & 0 & 0 & 0 & 0 & 0 & 0 & 0 & 0 & 0 & 0 \\
 & \labvc & 0 & 0 & 0 & 0 & 0 & 0 & 0 & 0 & 0 & 0 & 0 & 0 & 13.3 & 13.3 \\
 & \labboth & 0 & 0 & 0 & 0 & 0 & 0 & 0 & 0 & 0 & 0 & 0 & 0 & 0 & 0 \\
 \midrule
 \typeirr & \base & 6.7 & 6.7 & 6.7 & 0 & 0 & 0 & 0 & 0 & 6.7 & 6.7 & 0 & 6.7 & 6.7 & 0 \\
 & \labreg & 0 & 0 & 0 & 0 & 0 & 0 & 0 & 0 & 0 & 0 & 0 & 6.7 & 0 & 0 \\
 & \labvc & 0 & 0 & 0 & 0 & 0 & 0 & 0 & 0 & 0 & 0 & 0 & 0 & 0 & 0 \\
 & \labboth & 0 & 0 & 0 & 0 & 6.7 & 0 & 0 & 0 & 0 & 0 & 0 & 0 & 6.7 & 0 \\
 \midrule
\tokenreg & \base & 0 & 0 & 13.3 & 13.3 & 26.7 & 20.0 & 26.7 & 6.7 & 0 & 0 & 6.7 & 13.3 & 20.0 & 33.3 \\
 & \labreg & 0 & 0 & 20.0 & 0 & 13.3 & 13.3 & 6.7 & 0 & 13.3 & 0 & 0 & 6.7 & 0 & 6.7 \\
 & \labvc & 0 & 0 & 13.3 & 0 & 13.3 & 0 & 0 & 0 & 0 & 0 & 13.3 & 0 & 6.7 & 20.0 \\
 & \labboth & 0 & 0 & 20.0 & 13.3 & 0 & 0 & 0 & 0 & 0 & 0 & 0 & 0 & 0 & 6.7 \\
 \midrule
\tokenirr & \base & 13.3 & 13.3 & 40.0 & 13.3 & 40.0 & 6.7 & 26.7 & 0 & 13.3 & 6.7 & 26.7 & 6.7 & 33.3 & 33.3 \\
 & \labreg & 0 & 0 & 20.0 & 20.0 & 40.0 & 13.3 & 6.7 & 0 & 13.3 & 20.0 & 13.3 & 6.7 & 6.7 & 0 \\
 & \labvc & 6.7 & 0 & 20.0 & 20.0 & 0 & 6.7 & 0 & 0 & 13.3 & 0 & 0 & 0 & 13.3 & 6.7 \\
 & \labboth & 0 & 6.7 & 6.7 & 20.0 & 20.0 & 6.7 & 0 & 0 & 13.3 & 6.7 & 6.7 & 0 & 0 & 6.7\\
 \bottomrule
\end{tabular}
\caption{Test accuracy (\%) of different models on irregulars by verb class.}
\label{tab_app:vc_irregular}
\end{table*}

\subsection{Correlation}
The correlations with human data for different models are listed in Table~\ref{tab_app:cor}.
\begin{table*}[!h]
\small
\centering
\begin{tabular}{llllllll}
\toprule
&  & \multicolumn{2}{l}{Regular (N = 58)} & \multicolumn{2}{l}{Irregular (N = 58)} & \multicolumn{2}{l}{Irregular 2 (N = 16)} \\
\multicolumn{2}{l}{No Copy Mechanism} & $P_{pro} r$ & Rate $\rho$ & $P_{pro} r$ & Rate $\rho$  & $P_{pro} r$ & Rate $\rho$ \\
\midrule
\typereg & \base & 0.01 & 0.28 & 0.62 & 0.47 & 0.01 & 0.34 \\
 & \labreg & -0.14 & 0.23 & 0.33 & 0.28 & 0.20 & -0.02 \\
 & \labvc & -0.13 & 0.05 & 0.06 & -0.06 & 0.28 & 0.42 \\
 & \labboth & -0.04 & 0.28 & 0.43 & 0.17 & 0.23 & 0.14 \\
\typeirr & \base & -0.15 & 0.02 & 0.31 & 0.34 & NaN & NaN \\
 & \labreg & -0.02 & 0.46 & 0.36 & 0.28 & -0.23 & 0.01 \\
 & \labvc & -0.05 & 0.28 & 0.26 & 0.21 & 0.33 & 0.05 \\
 & \labboth & -0.02 & 0.48 & 0.40 & 0.37 & 0.56 & 0.13 \\
\tokenreg & \base & 0.57 & 0.51 & 0.29 & 0.33 & 0.33 & 0.11 \\
 & \labreg & 0.48 & 0.43 & 0.26 & 0.30 & -0.25 & 0.12 \\
 & \labvc & 0.57 & 0.59 & 0.22 & 0.22 & 0.12 & 0.36 \\
 & \labboth & 0.42 & 0.41 & 0.19 & 0.17 & -0.13 & 0.09 \\
\tokenirr & \base & 0.26 & 0.36 & 0.19 & 0.13 & 0.39 & 0.13 \\
 & \labreg & 0.24 & 0.41 & 0.23 & 0.22 & -0.16 & -0.02 \\
 & \labvc & 0.27 & 0.40 & 0.18 & 0.21 & -0.17 & 0.14 \\
 & \labboth & 0.30 & 0.43 & 0.20 & 0.14 & -0.05 & -0.04 \\
 \midrule
\multicolumn{8}{l}{Copy Mechanism} \\ 
\midrule
\typereg & \base & -0.14 & 0.20 & 0.12 & 0.30 & NaN & 0.45 \\
 & \labreg & -0.07 & 0.31 & 0.22 & 0.41 & NaN & NaN \\
 & \labvc & -0.11 & 0.28 & NaN & 0.44 & NaN & -0.03 \\
 & \labboth & -0.16 & 0.32 & NaN & 0.36 & NaN & 0.10 \\
\typeirr & \base & -0.04 & 0.36 & 0.29 & 0.51 & 0.64 & 0.08 \\
 & \labreg & -0.19 & 0.29 & 0.30 & 0.51 & NaN & -0.25 \\
 & \labvc & -0.12 & 0.33 & 0.23 & 0.55 & NaN & 0.14 \\
 & \labboth & -0.17 & 0.15 & NaN & 0.40 & NaN & NaN \\
\tokenreg & \base & 0.36 & 0.28 & 0.17 & 0.18 & 0.33 & 0.07 \\
 & \labreg & 0.35 & 0.32 & 0.23 & 0.21 & 0.26 & 0.08 \\
 & \labvc & 0.14 & 0.30 & 0.12 & 0.26 & -0.25 & -0.06 \\
 & \labboth & 0.18 & 0.16 & 0.13 & 0.08 & -0.04 & 0.05 \\
\tokenirr & \base & 0.30 & 0.27 & 0.26 & 0.30 & 0.29 & 0.09 \\
 & \labreg & 0.23 & 0.24 & 0.13 & 0.21 & -0.12 & 0.08 \\
 & \labvc & 0.27 & 0.32 & 0.16 & 0.22 & -0.25 & 0.04 \\
 & \labboth & 0.34 & 0.41 & 0.14 & 0.31 & 0.18 & 0.04 \\
 \midrule
\multicolumn{8}{l}{Resembing Methods Without Copy Mechanism} \\ 
 \midrule
\balance & \base & 0.62 & 0.74 & 0.44 & 0.45 & 0.69 & 0.28 \\
 & \labreg & 0.57 & 0.74 & 0.44 & 0.35 & 0.06 & 0.22 \\
 & \labvc & 0.63 & 0.70 & 0.47 & 0.42 & 0.42 & 0.33 \\
 & \labboth & 0.64 & 0.79 & 0.43 & 0.31 & 0.35 & 0.24 \\
 \midrule
\regds & \base & 0.61 & 0.74 & 0.46 & 0.51 & 0.55 & 0.11 \\
 & \labreg & 0.61 & 0.66 & 0.51 & 0.39 & 0.08 & 0.24 \\
 & \labvc & 0.48 & 0.60 & 0.42 & 0.40 & 0.49 & 0.51 \\
 & \labboth & 0.50 & 0.65 & 0.40 & 0.31 & 0.15 & 0.41 \\
 \midrule
\irrds & \base & 0.68 & 0.74 & 0.52 & 0.52 & 0.08 & 0.06 \\
 & \labreg & 0.52 & 0.63 & 0.39 & 0.33 & 0.77 & 0.43 \\
 & \labvc & 0.70 & 0.77 & 0.44 & 0.28 & 0.58 & 0.31 \\
 & \labboth & 0.49 & 0.63 & 0.39 & 0.34 & 0.39 & 0.09 \\
 \midrule
 \multicolumn{8}{l}{Resembing Methods With Copy Mechanism} \\ 
 \midrule
 \balance &\base & 0.54 & 0.70 & 0.34 & 0.32 & 0.48 & 0.11\\
 &\labreg & 0.51 & 0.69 & 0.49 & 0.39 & 0.42 & 0.40\\
 & \labvc & 0.65 & 0.75 & 0.31 & 0.23 & 0.30 & 0.36\\
 & \labboth & 0.52 & 0.63 & 0.45 & 0.42 & 0.50 & 0.30\\
 \midrule
 \regds &\base & 0.52 & 0.66 & 0.38 & 0.34 & 0.53 & 0.26\\
 &\labreg &0.54 & 0.67 & 0.37 & 0.35 & 0.28 & 0.34\\
 & \labvc & 0.50 & 0.69 & 0.49 & 0.43 & 0.43 & 0.31\\
 & \labboth & 0.51 & 0.67 & 0.31 & 0.21 & 0.40 & 0.15\\
 \midrule
\irrds &\base & 0.60 & 0.73 & 0.46 & 0.39 & 0.38 & 0.35\\
 &\labreg &0.63 & 0.76 & 0.42 & 0.34 & 0.56 & 0.48\\
 & \labvc & 0.64 & 0.71 & 0.34 & 0.20 & 0.46 & 0.18\\
 & \labboth & 0.59 & 0.67 & 0.38 & 0.31 & 0.62 & 0.16\\
 \bottomrule
\end{tabular}
\caption{Correlation with human data for different models. NaN represents the correlation that can not be computed due to too many zeros.}
\label{tab_app:cor}
\end{table*}

\end{document}